\documentclass[sn-mathphys]{sn-jnl}

\jyear{2021}

\theoremstyle{thmstyleone}

\theoremstyle{thmstyletwo}%

\theoremstyle{thmstylethree}%

\usepackage{lipsum}
\usepackage{booktabs}
\usepackage{subfig}
\usepackage[inline,shortlabels]{enumitem}

\begin{document}

\title[Iterative Self-Improved Model for Weakly Supervised Segmentation]{ISIM: Iterative Self-Improved Model for Weakly Supervised Segmentation}

\author*[1,2]{\fnm{Cenk} \sur{Bircanoglu}}\email{cenk.bircanoglu@adevinta.com} \email{cenk.bircanoglu@bahcesehir.edu.tr}

\author[2]{\fnm{Nafiz} \sur{Arica}}\email{nafiz.arica@eng.bau.edu.tr}
\equalcont{These authors contributed equally to this work.}

\affil*[1]{\orgdiv{Cognition}, \orgname{Adevinta}, \orgaddress{\street{24 rue des Jeûneurs}, \city{Paris}, \postcode{75002}, \country{France}}}

\affil[2]{\orgdiv{Computer Engineering Department}, \orgname{Bahcesehir University}, \orgaddress{\street{Besiktas}, \city{Istanbul}, \postcode{34353}, \country{Turkey}}}

\abstract{
Weakly Supervised Semantic Segmentation (WSSS) is a challenging task aiming to learn the segmentation labels from class-level labels. In the literature, exploiting the information obtained from Class Activation Maps (CAMs) is widely used for WSSS studies. However, as CAMs are obtained from a classification network, they are interested in the most discriminative parts of the objects, producing non-complete prior information for segmentation tasks. In this study, to obtain more coherent CAMs with segmentation labels, we propose a framework that employs an iterative approach in a modified encoder-decoder-based segmentation model, which simultaneously supports classification and segmentation tasks. As no ground-truth segmentation labels are given, the same model also generates the pseudo-segmentation labels with the help of dense Conditional Random Fields (dCRF). As a result, the proposed framework becomes an iterative self-improved model. The experiments performed with DeepLabv3 and UNet models show a significant gain on the Pascal VOC12 dataset, and the DeepLabv3 application increases the current state-of-the-art metric by \%2.5. The implementation associated with the experiments can be found: https://github.com/cenkbircanoglu/isim}.


\maketitle

\section{Introduction}\label{introduction}

In the last decade, researchers have achieved outstanding results on the semantic segmentation problem, one of the fundamental tasks of Computer Vision. These achievements have been mainly reached within the fully supervised setting where the training dataset contains pixel-level class labels. However, accessing that kind of dataset for a specific problem is a compelling task, and it is well-known that collecting pixel-level class labels is costly and time-consuming. Consequently, this issue diverted the attention of the researchers working on semantic segmentation tasks to apply another approach called Weakly Supervised Semantic Segmentation (WSSS), where weaker labels that are less costly to collect, such as class labels, bounding boxes, and scribbles, are employed to obtain pixel-level class labels.

Semantic Segmentation tasks in both fully supervised and weakly supervised settings have several challenges, such as cluttering of the background, occlusions, and intra-class variations. In addition to these challenges, there is another significant challenge in WSSS called the supervision gap. The supervision gap happens when the provided labels contain less information than the desired labels. For example, class labels contain information about the existence of the object classes in the image. Still, they do not hold any information about its shape, size, colour, or any other attributes. In addition, there is no knowledge of if one or multiple instances of a class exist in an image. Therefore, this gap significantly increases the complexity of the segmentation task and has a notable impact on substantial performance differences between WSSS and Fully Supervised Semantic Segmentation (FSSS) solutions.  

Most of the advanced studies in WSSS have employed a method called Class Activation Maps (CAMs) to obtain better segmentation labels in various ways. However, there is a significant issue with CAMs; they are interested in the most discriminative parts of the objects in the image. When CAMs are utilized to target segmentation labels, focusing on the most discriminating regions naturally concludes reaching non-complete and incorrect information about the objects and background. Most of the time, some parts of an object may be declared as background, or the other class instances may be missed in CAMs. Therefore, it is more reasonable to expect non-perfect results for the segmentation tasks with CAMs and accept that they require some modifications to reach more coherent results. 

The above observations make us focus on boosting the CAMs results with the implicit information pieces of the images to narrow the performance gap between FSSS and WSSS methods. In this study, we propose a general framework having an iterative approach to reveal these implicit information pieces of the images in a multi-task learning setting. The primary task of the framework is to train against the classification and segmentation labels simultaneously within the same network by using only the class labels and images as a source. The training objective against the segmentation label is scraping the implicit information about the objects from the images with the help of segmentation loss.

The proposed framework is built on an encoder-decoder-based segmentation model modified to support learning classification and segmentation simultaneously. As there are no ground truth values for the segmentation task, the same model also generates the segmentation labels. After giving the task of generation of the segmentation labels, fundamentally, the proposed framework becomes a self-improved encoder-decoder model supporting multi-task learning. Both encoder and decoder parts of the network are an instance of Fully Convolutional Networks (FCN). The overall network learns the semantic segmentation labels while the CAMs are produced as outcomes of training the encoder part and classification branch. The intuition behind this idea is that the encoder should carry the latent features of the objects to make the decoder learn segmentation labels while optimizing itself in class label training. 

The main objective of the proposed framework can be summarized as pixel-wise propagation of the CAMs by adding a pixel-wise loss to the training phase and using the image content. Additionally, to reinforce this propagation, dense Conditional Random Fields (denseCRF, dCRF) are employed in the production phase of pixel-level labels from CAMs.

There are three main contributions of this study, 
\begin{itemize}
    \item The proposed method is a general framework that can employ any encoder-decoder-based segmentation model aiming to achieve CAMs which is more coherent with segmentation results.
    \item To the best of our knowledge, it is the first iterative self-improved segmentation architecture with CAMs and dCRF.
    \item It is illustrated with the experiments on PASCAL VOC 2012 that our framework with a version of DeepLab achieves state-of-the-art performance with only image-level annotations.
\end{itemize}

The paper is organized as follows: after giving a detailed overview of related works on WSSS in section 2, the motivation of the study and proposed architecture are explained in the third section. Section 4 describes the implementation details, experimental results. Finally, Section 5 concludes the paper by discussing the proposed approach with experimental results. 

\section{Related Works}\label{related_works}

Segmentation is an essential ingredient of many systems built on image understanding. It plays a significant role in various applications from different domains such as autonomous driving, analysis of biomedical images, video surveillance, augmented reality, and fashion. Image segmentation algorithms designed to support these applications have led to numerous methods for an extensive history of image segmentation. Even though there are many studies in the literature using traditional Computer Vision and Machine Learning methods, such as thresholding \cite{4310076}, clustering \cite{DHANACHANDRA2015764}, Conditional and Markov Random Fields \cite{10.1145/1553374.1553479}, and others \cite{Neubert2014CompactWA, Kass2004SnakesAC, 10.1007/978-3-540-88693-8_52}, our focus in this study is on the latter approaches that apply Deep Learning methods.

In recent years, FSSS applications have become state-of-the-art on the popular benchmark datasets using Deep Learning methods \cite{setr, wu2020cgnet, kirillov2020pointrend, yin2020disentangled, He_2019_ICCV, He_2019_CVPR, deeplabv3plus2018}. The main objective of FSSS is assigning the correct class label to each image pixel by observing the pixel-level ground truths. To fulfil this objective, the authors \cite{DBLP:journals/corr/LongSD14} introduced the Fully Convolutional Network (FCN) to obtain low-dimensional segmentation masks by converting the fully connected layers to convolutional layers in CNN architecture. Conditional Random Fields (CRFs) were added as a post-processing step by executing it on the results of FCN to enhance their performances \cite{chen2016semantic}. Noh et al. \citep{DBLP:journals/corr/NohHH15} proposed a new model which consists of two parts, an encoder and a decoder, to obtain relatively high-dimensional segmentation masks. In the network, an instance of FCN takes charge of the encoding part, and the decoder part employs the transposed convolution layers. This study enlarges the dimension of the segmentation result without using interpolation directly on the output of an FCN. One representative example of the encoder-decoder-based method called UNet \cite{unet} employed specific connections between the encoder and decoder blocks and symmetric expanding paths. In addition to these, there are two popular families in FSSS called DeepLab \cite{deeplab, deeplabv3, deeplabv3plus2018} and R-CNN based models \cite{DBLP:journals/corr/HeGDG17}. DeepLab introduced the usage of dilated convolutions in FCN. R-CNN-based models introduced Region Proposal Network, which proposes the candidate regions and extracts the Region of Interest (RoI) and RoIPool layer, which computes the features from the proposals. There are also other methods, which are popular because of their performance, elegance, or simplicity \cite{DBLP:journals/corr/LinDGHHB16, DBLP:journals/corr/ZhaoSQWJ16, DBLP:journals/corr/GhiasiF16, DBLP:journals/corr/VisinKCBMC15, DBLP:journals/corr/ChenYWXY15}.

Likewise, the objective in WSSS is the designation of each pixel of the images with class labels. However, WSSS applications aim to fulfil this objective by learning from image-level ground truths instead of pixel-level ground truths. This critical difference significantly affects the literature on WSSS as it requires applying different techniques than FSSS.

In recent years, various studies have been published on achieving segmentation labels by employing weaker labels, and these studies can be broadly categorized into four main approaches  \cite{DBLP:journals/corr/abs-1912-11186}; 
\begin{itemize}
    \item Expectation-Maximization (EM)
    \item Multiple Instance Learning (MIL)
    \item Self-Supervised Learning (SSL)
    \item Object Proposal Class Inference (OPCI)
\end{itemize}

In the following subsections, we introduce the above categories by giving the details of decisive studies in each of them. However, the studies on natural scene images take more of our attention even though there are WSSS applications \cite{DBLP:journals/corr/abs-1808-04277, 9009552, 8953780, DBLP:journals/corr/LinCDWQH17, DBLP:journals/corr/abs-1904-03983, robinson2019large, DBLP:journals/corr/abs-1806-03510, Wang2018WeaklySL} in other domains such as histopathology and satellite images. They are not considered in scope due to the differences in the characteristics of the solutions and datasets.

\subsection{Expectation-Maximization}
\label{expectation_maximization}

Theoretically, Expectation-Maximization is an iterative approach that contains two main tasks: optimizing a latent distribution across the image and learning the segmentation masks from that latent distribution. In practice, the researchers follow these steps, 
\begin{enumerate}
    \item Generate the segmentation labels from images and class labels with a prior assumption
    \item Train a Fully Convolutional Network to learn segmentation labels
    \item Regenerate the segmentation labels from images, class labels, and the features learned by FCN
    \item Iterate over steps 2 and 3
\end{enumerate}
One of the early studies of this category, called CCNN \cite{DBLP:journals/corr/PathakKD15},  converted the original problem into a biconvex optimization problem and solved it by optimizing the convex latent distribution of fixed FCN outputs with several constraints while training the FCN against the fixed latent distribution. One representative study for this category called EM-Adapt \cite{DBLP:journals/corr/PapandreouCMY15} sets the expectation by adding a pixel-level bias to the FCN according to the given class labels. In the maximization phase, it handles the expectation by optimizing the outputs of the FCN to target these pixel-level labels.

\subsection{Multiple Instance Learning}
\label{multiple_instance_learning}

Multiple Instance Learning (MIL) is one of the approaches of the supervised learning framework that deals with learning from labels containing incomplete information. In practice, this method solves the WSSS task by having a model that aims to learn the class labels of the image and then assign each pixel with one of the given class labels for a given image and its class labels. In general, the WSSS application of MIL comes across as the training of a Convolutional Neural Network (CNN) with image-level loss and inferring the image locations according to the class-level prediction. Class Activation Map (CAM), which is extensively used in the literature for several reasons and ways by other studies like this study, is an early approach of the attention mechanism on CNN. Mainly, it was proposed to understand, explain, and explore CNN architectures better by achieving the activated areas of the image according to the model’s predictions. In CAM study \cite{cam}, there are two constraints, (i) CNN architecture should be an instance of a Fully Convolutional Network (FCN) (ii) There should be a Global Pooling layer before the classifier layer, which is a convolutional layer. Within these constraints, the calculation of the CAMs can be accessed by the dot product between the last two convolutional layers. As the calculation of CAMs requires some modifications in the well-known CNN architectures, the training step is mandatory for this initial study. There are other studies to extend the work of CAMs to improve the results or ease the process. For example, in Grad-CAM \cite{DBLP:journals/corr/SelvarajuDVCPB16}, the authors used guided backpropagation to reach CAMs without changing any part of the original network and training the network. WILDCAT \cite{8100114} offered a more complex pooling layer to enrich the activated areas by following mainly the idea of the original CAM study \cite{cam}. The authors \cite{pathak2015fully} converted the VGG-16 \cite{vgg} network to an FCN by replacing the Dense layers with a 1x1 Convolutional layer also employed Global Average Pooling as the pooling layer and trained the modified network to learn class-level annotations with a special MIL loss offered. In the end, the top predictions at each location were upsampled bilinearly to achieve the segmentation results. The authors  \cite{10.1007/978-3-319-46493-0_14} utilized a specific method called guided backpropagation to reach the pixel-level top-class predictions. First, with the help of guided backpropagation, authors created the coarse class activations for multiple convolution layers of the network and aggregated them after unifying them into one scale. Additionally, the process ended with the employment of denseCRF. The authors \cite{DBLP:journals/corr/SalehASPGA16} trained an FCN by targeting the class labels to generate prior foreground/background masks from the intermediate convolution layers; then, the authors used these masks to learn segmentation masks. 

\subsection{Self-Supervised Learning}
\label{self_supervised_learning}

Lately, Self-Supervised Learning has become the most popular approach with its diverse and performant examples for WSSS. The applications in this category mostly contain two steps; one to solve pretext tasks to use the outputs of it in the other task, the other one is a more complex downstream task which is most of the time the actual task. The pipeline often starts with learning pseudo semantic segmentation labels from the given class labels with one network and continues with a segmentation model to target the generated pseudo labels. In most studies, the first network aims to produce CAMs, and then the obtained CAMs are used in the training phase of the segmentation network.

In the literature, there are a vast amount of examples in this category as they reach good performances. SEC \cite{DBLP:journals/corr/KolesnikovL16}, which can be described as the best matching study to the given pipeline above, trained a VGG-16 on class labels to produce pseudo-pixel-level labels with the outcome of CAM and denseCRF \cite{DBLP:journals/corr/abs-1210-5644}. After the training of the first model, DeepLab-LargeFOV \cite{chen2016semantic} was employed against the produced labels with a constraint loss as a segmentation model. Another study called MDC \cite{DBLP:journals/corr/abs-1805-04574} followed the same steps with SEC with one modification; they assigned multi-dilated convolutional layers to the pseudo-ground-truths and generated the CAMs from these multi-dilated layers. AE-PSL \cite{DBLP:journals/corr/WeiFLCZY17} is another study that follows SEC’s steps; with a significant novelty and a better performance, an adversarial erasing mechanism was integrated into the model to erase the activated regions of the previous steps to encourage the network to learn less discriminative parts. FickleNet \cite{DBLP:journals/corr/abs-1902-10421} employed Grad-CAM instead of CAM to achieve a better performer segmentation model and added a centre-fixed spatial dropout to the later convolutional layers to produce the pseudo-ground-truth labels to train an FCN against them. However, Grad-CAM also suffered from the same issue as it focuses on discriminative regions of the objects and the inputs of the second model are not precise. 

Later, the researchers started to focus on how to take the CAM results as seed points to propagate from them. The idea behind this is that, hypothetically, adjacent pixels of the activated ones have more possibility to be a part of the same object than the others. Therefore, propagating from CAMs to the adjacent regions can efficiently produce better pseudo-ground-truths. With this in mind, the authors of DSRG \cite{huang2018dsrg} proposed a region-growing approach from CAMs to produce pixel-level labels. There is one important study in the literature called Pixel-level Semantic Affinity (PSA) \cite{DBLP:journals/corr/abs-1803-10464}, which dramatically influences other studies, including ours, with its proposed architecture. The pipeline of PSA has three steps to follow and starts with CNN training to generate class activations as in SEC. In the second step of the pipeline, instead of using the CAMs as ground-truth segmentation labels, the CAMs were used as seed points, and one more model was added to the pipeline, which performs a random walk to propagate from the seeds to achieve the pseudo-ground-truth for training a segmentation model. IRNet \cite{DBLP:journals/corr/abs-1904-05044} can be called a version of PSA, which also targets a more complex task called instance-level semantic segmentation, with additional pixel-wise clustering to the random walk process. It performs the random walk from low-displacement field centroids in the CAM seeds until the class boundaries to produce pseudo-ground truths. Recently there are more advanced studies influenced by PSA were published. One of them, PuzzleCAM \cite{jo2021puzzlecam}, significantly improved overall performance by obtaining CAMs on image patches. The network to obtain the CAMs transformed to a siamese network without changing its aim. One branch of this siamese network was kept the same, but the other was organized to create tiles from the input image and merge the CAMs of these tiles. In addition to these, a reconstruction loss was proposed to regulate the results of both branches. There are also some studies to improve CAMs using saliency maps as additional information, such as  \cite{DBLP:journals/corr/abs-2105-08965, DBLP:journals/corr/abs-1910-05475}. Even though most of the studies focus on improving the first model, CAMs, in the pipeline offered in PSA, one study \cite{DBLP:journals/corr/abs-2103-16762} focused on the feature propagation frameworks and examined the Graph Convolutional Network (GCN) instead of AffinityNet to propagate from CAMs. To expand object activation regions like the others, DRS \cite{DBLP:journals/corr/abs-2103-07246} used a different approach than others, and it suppressed the attention on discriminative regions and spread the attention to adjacent non-discriminative regions by generating dense localization maps.

\subsection{Object Proposal Class Inference}
\label{object_proposal_class_inference}

Most of the applications in this category first employ a low-level feature extractor, mainly with one of the Traditional Computer Vision methods, and use the features to generate the pseudo-ground-truths. In the applications, researchers assemble methods from MIL and SSL to solve the WSSS problem. The authors \cite{Kwak2017WeaklySS} aggregate the superpixels and the features extracted from a CNN to generate the pseudo-ground-truths to train an FCN against them. In PRM \cite{DBLP:journals/corr/abs-1804-00880}, the low-level object proposals were secured by employing Multi-scale Combinatorial Grouping (MCG) \cite{DBLP:journals/corr/Pont-TusetABMM15}, and an FCN was trained with the loss they proposed called peak stimulation loss. As the next step, they applied peak backpropagation to convert Class Response Map to Peak Response Map for each peak. By performing non-maximum suppression on the class labels, the object proposal extracted from PRM peaks. SPML \cite{DBLP:journals/corr/abs-2105-00957} employed a Contrastive Learning approach with its four types of relationships between pixels and segments in the feature space, capturing low-level image similarity, semantic annotation, co-occurrence, and feature affinity to access the segmentation masks.



\section{Methodology}\label{methodology}

In most WSSS approaches, CAMs are used to obtain pseudo-segmentation labels as it is or as prior information for the subsequent stages. We also utilize CAMs \cite{cam} and propose a generic framework, which iteratively improves the semantic segmentation model in an encoder-decoder-based architecture. This new framework contains elements from all the WSSS categories mentioned in section \ref{related_works}. The proposed method is an example of EM due to its iterative approach, and the overall pipeline is an SSL application. To generate the pseudo-segmentation labels, CAMs, which is an application of the MIL approach, are employed. As CAMs are improved with the help of dCRF, it also contains the elements from OPCI. 

In this section, after revealing our motivation, the proposed framework is described in detail.

\subsection{Motivation}\label{motivation}

The utilization of CAMs to attain pseudo-segmentation labels is highly popular in the literature and is widely used in WSSS applications. These applications achieved relatively great results on the benchmark datasets; however, they have two main issues that can be improved from our perspective. 

The first issue appears due to the standard process of CAMs generation. The typical procedure of obtaining CAMs contains a classification model optimized with a classification loss that mainly focuses on the most discriminative parts of the objects in the image that, causes to ignore the less discriminating parts. Ignoring less discriminative regions or concentrating on some parts of the objects is reasonable for a classification model as its primary interest is the information of the object's existence in the image. However, when it comes to the segmentation models, every single pixel is equally essential. As a result of this contradiction between the nature of classification and segmentation, employing standard CAMs procedure to generate pseudo-segmentation labels in WSSS applications can perform well; however, it can not achieve the optimal solution by itself. In light of this information, injecting a pixel-level loss to the training phase may help to achieve more coherent CAMs hypothetically by trying to equalize the importance of the pixels and learning from the image content. This pixel-level loss will help activate the less discriminative regions of the objects in the CAMs. 

The other downside of using CAMs appears in the transition step of creating the pseudo-segmentation labels from them by applying a threshold. This thresholding process on the obtained CAMs is vital as the decision about each pixel belonging to an object or background is made here. This transition process mainly results in non-complete, non-precise results. The most identified problem is the wrong categorization of a pixel of the foreground object as a background pixel. Therefore, working on these pseudo-segmentation labels as ground truth significantly affects the next steps. Moreover, in previous applications, there is no link between the threshold value and the training steps of the classification network. Therefore, we propose an iterative process with a pixel-level loss by setting this threshold at the beginning of the training and implicitly optimizing CAMs and the classification network according to this threshold value.

These observations above bring us to the point that the CAM-based WSSS model should aim to learn from the whole body of objects and be boosted with a generated pixel-level loss. We think that it can be achieved by iteratively improving the CAMs with the help of a pixel-level segmentation loss using pseudo-ground-truths at each step of the training phase. In addition, by deciding the threshold value at the beginning of the process and optimizing the model, CAMs can be crucial to have more coherent pseudo-segmentation labels. With this motivation, we propose a framework based on an encoder-decoder segmentation model that generates its segmentation labels from the classification branch with the help of denseCRF and thresholding process and learns these labels with the decoder part while learning the classification labels simultaneously and iterating over this process multiple times to improve itself at each iteration.

\subsection{Proposed Framework}\label{proposed_framework}

We follow the general pipeline of the PSA method \cite{DBLP:journals/corr/abs-1803-10464} previously proposed and used by many other studies such as PuzzleCAM, CDA, and SEAM \cite{jo2021puzzlecam, cda, seam}. This section first presents the PSA pipeline and then describes the modifications in the pipeline as a proposed framework in detail. 

\begin{figure}[H]
    \centering
    \includegraphics[width=1.0\columnwidth]{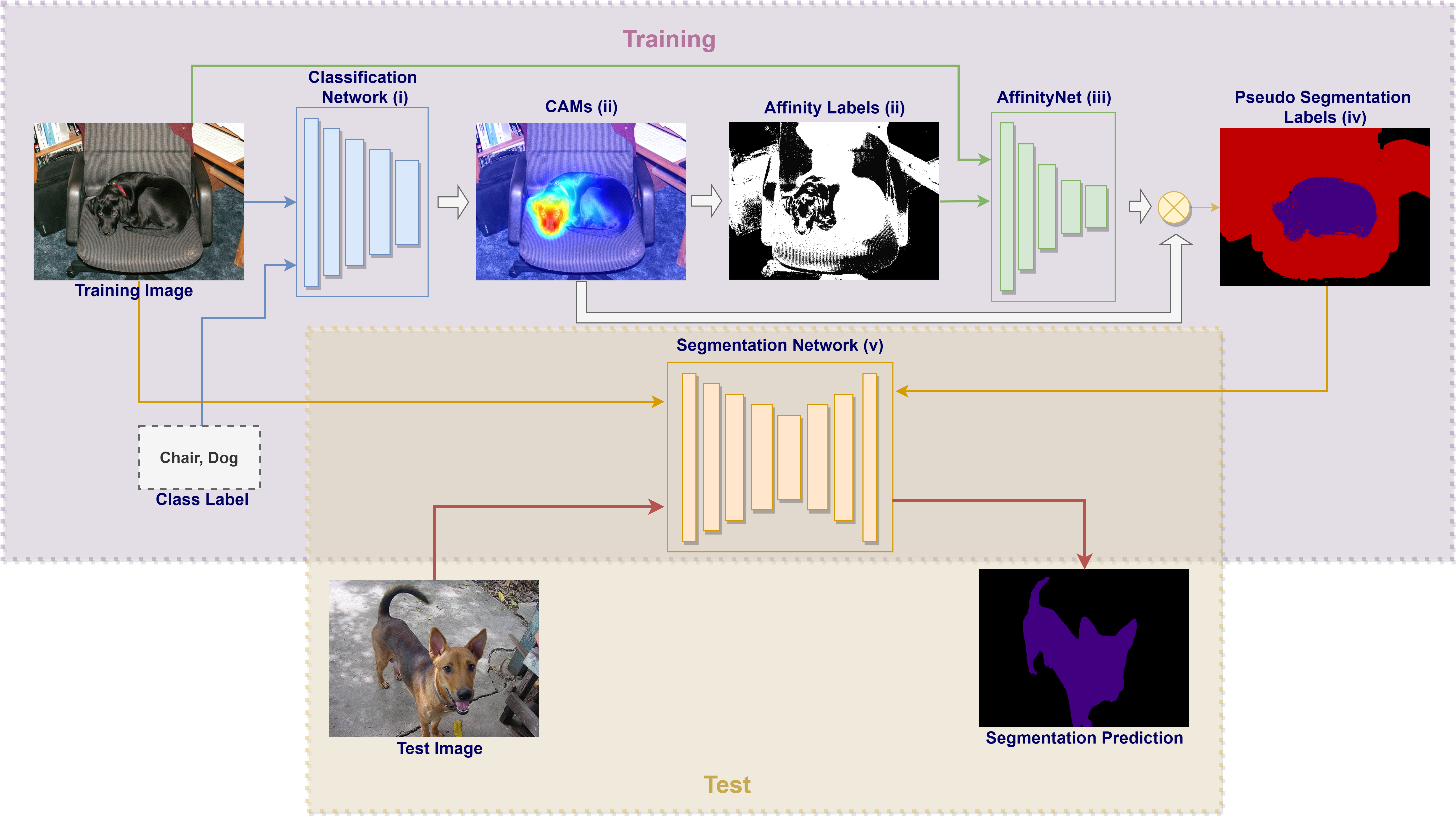}
    \caption{End-to-End Pipeline of PSA}
    \label{end2end_pipeline}
\end{figure}

The PSA pipeline contains several steps, visualized in Figure \ref{end2end_pipeline} in the training phase. It is a representative application of a two-stage framework with a Self-Supervised Learning approach to achieve a semantic segmentation model using class labels. The two stages of Self-Supervised approaches are the generation of the pseudo-ground-truths in the first stage and targeting them in the second stage. In the first stage, after obtaining the CAMs, PSA offers a new model called AffinityNet, which propagates CAMs by using class-agnostic similarities to generate segmentation labels. The segmentation network is trained in the second stage by taking these labels as pseudo-ground-truths. Briefly, PSA contains five steps in the training phase to fulfil this two-stage framework: (i) training of the classification model to obtain CAMs, (ii) generation of pixel-level affinity labels from CAMs with applying a threshold and denseCRF, (iii) training AffinityNet to learn class agnostic pixel-similarities targeting pixel-level affinity labels, (iv) generation of the segmentation labels from the combination of CAMs and predictions of AffinityNet, (v) training of the segmentation model using segmentation labels. The classification model initially takes the training images with their class labels and produces CAMs. Then these CAMs are processed with threshold and denseCRF to reach affinity labels. AffinityNet is trained with the images and their affinity labels. Then segmentation labels are generated by applying a random walk algorithm on CAMs based on the affinity matrix produced from AffinityNet. Finally, the segmentation network is trained by taking those segmentation labels as pseudo-ground-truths. Segmentation Network is used in the test phase to obtain segmentation results in the test images.

In PSA-based approaches, adjustments are primarily made in the first step to improve its overall performance by reaching more accurate CAMs and affinity labels. As the problem of the CAMs mainly comes from the loss function and employing a classification network to obtain them, the most straightforward alternative is employing a segmentation network with a segmentation loss. However, as there are no ground truths for the segmentation task, this requires extra supplementary updates. 

This study proposes an iterative approach to train a model that concurrently targets both the classification and segmentation tasks while generating pseudo-segmentation labels. The proposed model is an encoder-decoder-based segmentation model, which enhances the pseudo-segmentation labels and CAMs at each iteration.

\begin{figure}[H]
    \centering
    \includegraphics[width=1.0\columnwidth]{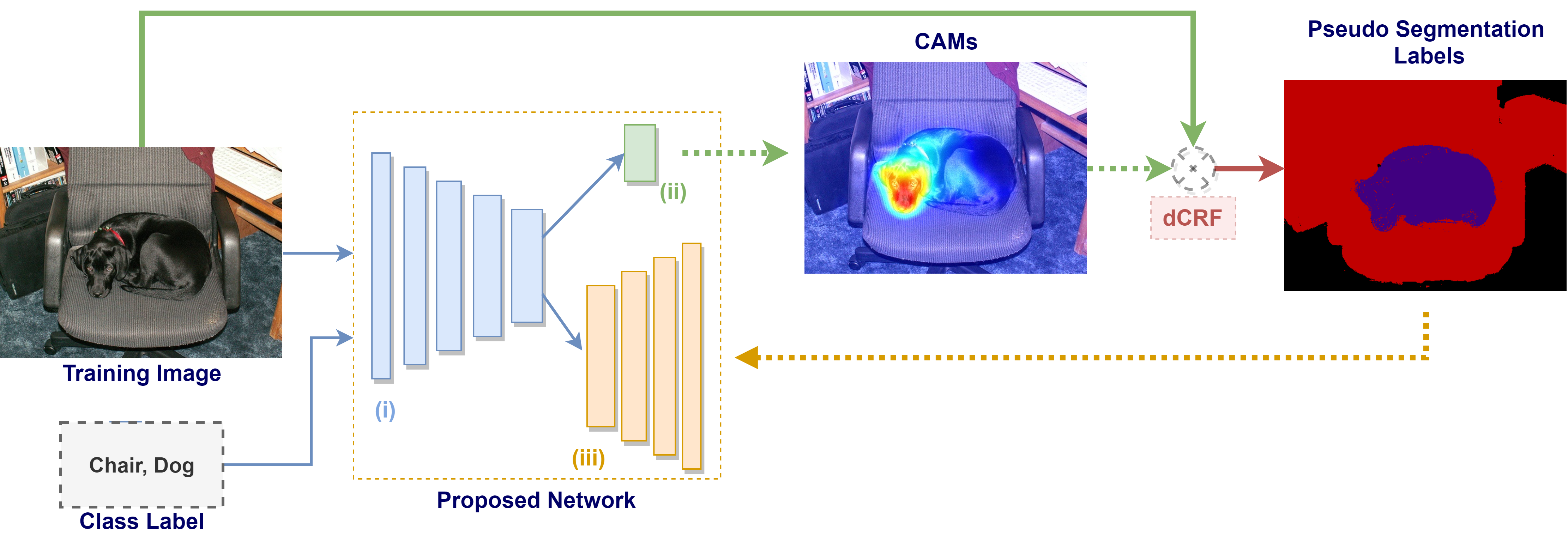}
    \caption[Proposed Architecture]{Proposed Architecture: (i) Encoder network, (ii) Classification Branch, (iii) Decoder network}
    \label{proposed_architecture}
\end{figure}

In the proposed encoder-decoder-based architecture, different parts of the model share the tasks for WSSS learning. The encoder network learns the class-level annotations while discovering the latent features in CAMs to support the segmentation task. The decoder network aims the pixel-level annotations. The training steps of the proposed architecture are given in the pseudo-code Algorithm-\ref{algorithm} and visualized in Figure \ref{proposed_architecture}. An overall training phase is an iterative approach containing three main steps. Even though it can be divided into three steps to make it easy to explain and formalize, in practice, the training is done in one step without stopping or starting it. Given the training set of images and their class labels, the first step is to train the encoder network, from which we can extract the CAM of each training image. This step can be considered the initialization of the whole model. In the second step, pseudo-segmentation labels are produced with the help of the denseCRF algorithm. In the last step, the whole model is retrained by taking the pseudo-segmentation labels as ground truth. Two different methods are employed for the pseudo-segmentation label generation and encoder-decoder training steps. In the first one, iteration continues until convergence, and in the second one, fixed-size iterations are used. The performances of these two methods are given in the experimental results.

\begin{algorithm}
\caption{Training steps of the Proposed Architecture}\label{algorithm}
\begin{algorithmic}[1]
\Procedure{Train}{$images, class\_labels$}
\State $x\gets images$
\State $z\gets class\_labels$
\State $y\gets null$
\State $Model\gets Segmentation Model with Pretrained ImageNet$
\State $Model$ = \texttt{TrainEncoder($Model$, $x, z$) } 
\For{\texttt{i = 1 to N}}
    \State y = \texttt{GeneratePseudoGroundTruths($Model$, $x, z$)}  
    \State $Model$ = \texttt{TrainEncoderDecoder($Model$, $x, y, z$)} 
\EndFor
\State \Return {$Model$}
\EndProcedure
\end{algorithmic}
\end{algorithm}

%
%
%


The only architectural change in the encoder-decoder-based segmentation model is that a new classification branch, marked as (ii) in Figure \ref{proposed_architecture}, is added after the encoder network with Global Average Pooling (GAP) and \((1x1)\) Convolutional layer to support classification and to produce CAMs. The proposed architecture is generic to practice with any encoder-decoder-based segmentation model.

In the training procedure, we utilize denseCRF to refine CAMs and practice the outputs as pseudo-ground-truths for the segmentation task. Utilizing denseCRF over CAMs and accessing the segmentation labels have two implicit benefits. First, to improve the CAMs, the power of denseCRF is embedded in the encoder network. Second, the information pieces of the images become more extractable for the model. After generating and using segmentation labels in the model, the problem becomes multi-task learning. With these changes, the encoder network is simultaneously optimized to learn the discriminative regions and the other regions related to the object with the help of classification and segmentation loss, respectively. Hypothetically, learning other features should positively affect the CAMs, and to challenge this hypothesis, we extensively operate experiments on the proposed framework.

However, in practice, the proposed framework has two potential issues which can affect the experiments significantly and make the training unstable. The first one is observed when each pixel of the pseudo-segmentation mask is marked as the background. The second issue is that classification or segmentation loss suppresses the other one; consequently, both or one of them starts to diverge. The first issue is suspected to be seen mostly when the classification network is not confident enough to distinguish between the classes. In addition, discriminating between the classes is not enough as a threshold is applied to CAMs. The model should have a higher confidence score in the relevant pixels than the threshold value. Also, it is reported in Section \ref{experiments} that the classification models are not 100\% accurate on predictions. Therefore, there is a possibility of inconsistency in the pseudo-ground-truths, affecting the overall training process. The decoder part of the model becomes confused when there are all background pixels as targets for the segmentation task. As a result, it will affect the encoder part and may make the classification loss diverge.  
The function given in Eq \ref{modified_loss} is proposed to ease this issue. The aim of the proposed function is that when there is a segmentation mask that involves only background pixels, ignore that mask and do not calculate any loss for that specific image.

Supposed that \(y_i\) is the threshold applied CAM for the instance \(i\) and it is defined as,

\begin{equation}
y_i = 
 \begin{pmatrix}
  a_{1,1} & a_{1,2} & \cdots & a_{1,n} \\
  a_{2,1} & a_{2,2} & \cdots & a_{2,n} \\
  \vdots  & \vdots  & \ddots & \vdots  \\
  a_{m,1} & a_{m,2} & \cdots & a_{m,n} 
 \end{pmatrix}
 \label{eq_y_i}
\end{equation}



the formulation of the segmentation loss for one instance becomes as below with our basic modifications,
\begin{equation}
\text{ModifiedCrossEntropyLoss}(\hat{y}, y) = \text{CrossEntropyLoss}(\hat{y} \cdot r_y, y)
\label{modified_loss}
\end{equation}
where \(r_y\) is defined as ,

\begin{equation}
r_y  =  \begin{cases}
                0, & \text{if } \sum_{j=0,k=0} a_{jk} = 0 \\
                1, & \text{else} \sum_{j=0,k=0} a_{jk} > 0
 \end{cases}
 \label{eq2}
\end{equation}

On the second issue, instead of generating the pseudo-ground-truths in each epoch, we create them with a different frequency to give some room to the network to learn class and segmentation labels. On this point, we investigate and report the outcomes in the following sections.

To challenge the proposed framework and process, we concentrate on two common and well-known segmentation models, one from the DeepLab family, DeepLabv3, and the other is the UNet model. The original architectures of DeepLabv3 and UNet are modified according to the recommended method. The modified architectures are called DeepLabCAM and UNetCAM referred to in our study.

\subsubsection{DeepLabCAM}

\begin{figure}[H]
    \centering
    \includegraphics[width=1.0\columnwidth]{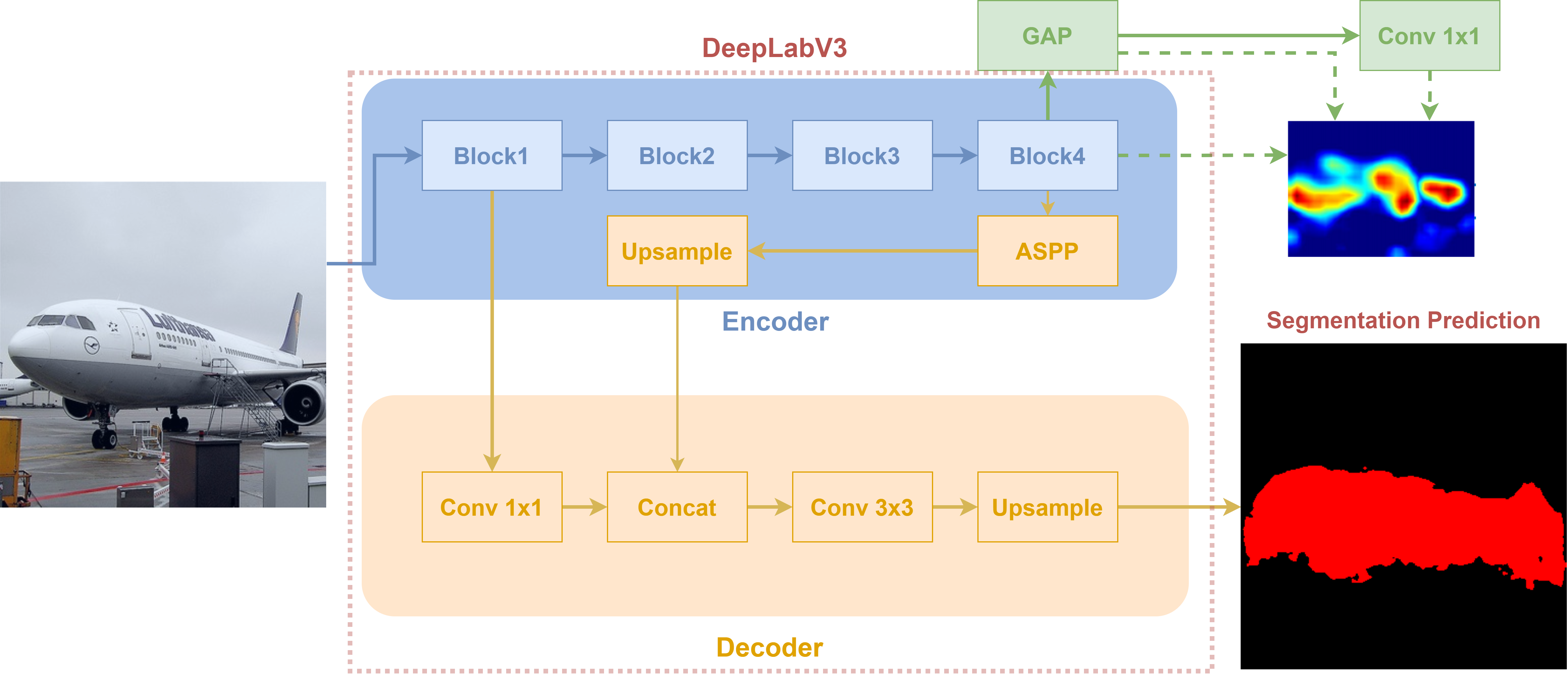}
    \caption{DeepLabCAM Architecture}
    \label{deeplabcam_architecture}
\end{figure}

DeepLab is one of the most popular image segmentation approaches in the literature. The networks belonging to this family achieved marvellous results on various tasks in fully supervised settings. In this study, one version of DeepLab called DeepLabV3 is preferred as a segmentation network to transform for a strong reason as the primary network. The literature shows that the dimension of the CAMs has an essential role in the results. Most of the applications in the literature adopt a network from the ResNet family to obtain the CAMs. When the original ResNet is applied for that task, the output dimension of the CAMs becomes \(16x16\), and with experiments, it is shown that using \(32x32\) dimensioned CAMs perform better.
Moreover, the dimension is doubled by changing the dilation rate of the last block of the ResNet family. In addition, one of the critical features of the DeepLab family is denoted as using the dilated convolutional layers. With these in mind, we add the classification branch after the fourth convolutional block of the encoder part of the DeepLabv3, as shown in  Figure \ref{deeplabcam_architecture}.

\subsubsection{UNetCAM}\label{unetcam_description}

UNet is also one of the most popular architectures in segmentation studies with unique characteristics and features. Two different forms of the UNet are modified and executed in the experiments to prove the hypothesis that adding segmentation loss affects the CAMs positively.

In both UNetCam versions, ResNet50 is employed as the encoder part, and the decoder part is implemented concerning the constraints in the original UNet architecture. In UNet, there are skip connections between the encoder and the decoder blocks, and these skip connections are kept as they are. The difference between the UNetCAM versions comes from the implementation details of the upsampling layers. The first version contains non-learning upsampling layers, and the second one has learnable deconvolution layers. Also, the classification branch is injected after the encoder network in both networks, as proposed.

The implementation details and the experimentation settings are given in the following section. The experimentation settings are organized to challenge the proposed framework from multiple perspectives, such as genericity, stability, and performance. 



\section{Experiments}\label{experiments}

The proposed framework is assessed under numerous settings from four different perspectives. First of all, to reveal its primary effect on the performance metric, the proposed framework is compared extensively with its original form, CAMs. As the second, two different segmentation models are transformed according to the proposed method, and they are executed to support the genericity of the approach. As a third perspective, we study the effects of the pseudo-segmentation regeneration frequency as the proposed training procedure possesses iterations over the network. The fourth collection of experiments is arranged related to the impact of the size of the images, as image resolution has an apparent effect on the segmentation part.

After investigations introduced above are concluded on the proposed method, another experimentation set is practised to challenge the other conjecture improving CAMs is the key to enhancing WSSS results. In that direction, the pipeline mentioned previously is applied by employing the CAM results of our proposed framework. The following parts of the section give all the details of the experiments and the results.

Additionally, another comparison is made between the end-to-end pipeline with the proposed framework and the state-of-the-art methods in the literature.

All the experiments are performed on PASCAL VOC 2012 dataset \cite{Everingham10} with 20 class annotations and additional background annotation. The splits of the official dataset have 1464, 1449, and 1456 images for training, validation, and testing, respectively. We obey the conventional experimental protocol to make fair comparisons and take the additional annotations from SBD \cite{BharathICCV2011} to build an augmented training set. There are 10582 images in this extended training set, and it is named the trainaug set in this study. In addition to the images, we only use image-level classification labels during training phases. 

Furthermore, segmentation results are assessed on the mean intersection over union (mIoU) metric. In addition to metrics, visualizations of the pseudo-segmentation labels are represented to investigate the effects of the proposed method. The dataset does not contain the ground-truth values for the test split, and to evaluate our results, we use the evaluation server maintained by the owners of the dataset. This evaluation server calculates the test split metrics, which are used as it is. Even though there are ground-truth values for the validation dataset, the same method is preferred among the validation split, aiming for consistency in the results.

Additionally, all the experiments are implemented in PyTorch \cite{NEURIPS2019_9015}, and a computer with 4 RTX 2080 GPU is used to operate the end-to-end pipeline.

\subsection{Experimentations on the Proposed Framework}

The proposed framework is challenged from several perspectives. Before presenting the experimental results, we introduce all the crucial functions and parameters applied in these experiments by grouping them as preprocessing, generating pseudo-ground-truths, and training.

\textbf{Preprocessing:} In the training phase, the images are resized according to these rules as follows,
\begin{itemize}
    \item If the longer side of the image is larger than 640, it is scaled down to 640.
    \item If the longer side of the image is smaller than 320, it is scaled up to 320.
\end{itemize}

Also, random horizontal flip and random cropping are applied to the images with the given resolution value of 320×320 if not otherwise declared, and images are normalized between 0 and 1. Also, the exact same cropping and scaling operations are followed for the pseudo-segmentation labels. In the CAMs prediction phase, four different scaled versions and their horizontally flipped images are used separately to generate CAMs, and the outputs are concatenated into one. 1, 0.5, 1.5, and 2 values are used as scales. 

Although there are possible other data augmentation options, we employ the same steps as the previous studies to make the comparison fair. 

\textbf{Generation of Pseudo-ground-truths:} Three steps are applied to generate pseudo-ground-truths, respectively (i) Producing CAMs as a result of encoder and classification branch, (ii) Applying a threshold value to decide which pixel involves foreground and background, and (iii) Executing dCRF on the thresholded CAMs. The threshold value is chosen as 0.3, although choosing another value for this threshold does not have a significant effect if they are not too big or too small, as the network feeds itself with the outcomes and reaches the balance with the result of the iterations. Also, this behaviour can be pointed out as a significant difference from the previous studies as it is a new addition to the CAM process. Additionally, Unary potentials are used in dCRF calculations.

\textbf{Training Process:} To support the classification and the segmentation simultaneously, the model employs two different loss functions. The multi-label classification task calculates the loss with the Multi-Label Soft Margin loss function, and modified pixel-wise Cross-Entropy loss, which is described in Section \ref{proposed_framework}, is adopted for segmentation tasks. Additionally, Poly optimizer, an extension of Stochastic Gradient Descent (SGD), is employed with momentum value 0.9 with different initial learning rates for the encoder, decoder, and classifier layers as 0.1, 0.01, and 1, respectively. 

The training is performed for 50 epochs in the default settings, and the first five epochs are only trained among classification labels. After the fifth epoch, the network is trained with both class and segmentation labels for each epoch. In default settings, segmentation labels are regenerated every ten epochs after the fifth epoch.  

\subsubsection{CAM vs. DeepLabCAM}

An extensive set of experimental configurations is designed to prove our hypothesis and evaluate our framework. First of all, to perceive the effects of the proposed method, several variations of it and the corresponding implementation of CAM formed. In this direction, seven different backbones are employed to compare CAM and DeepLabCAM methods. 

In the first experiment set, the classifier backbones of the DeepLabCAM and CAM are adjusted as the versions of the ResNet and ResNeSt network family. Dilated convolution is used in the last blocks of all networks to double the dimension of the CAMs.

Before presenting the segmentation results, classification scores are shared in Table \ref{classification_scores}. Even though previous WSSS studies ignored these metrics, they are essential metrics as the accuracy of the classification model affects the CAM performance directly. If the model does not correctly predict the class, there will be no information about it to learn in the following steps of the pipeline. As shown from Table \ref{classification_scores}, models containing ResNeSt-based backbones perform better in the classification task and, most probably, obtain better CAMs.

\begin{table}[!ht]
    \begin{center}
        \begin{minipage}{\textwidth}
            \caption{Classification Accuracy}\label{classification_scores}%
            \begin{tabular*}{\textwidth}{@{\extracolsep{\fill}}cccccccc@{\extracolsep{\fill}}}
                \toprule
                & \multicolumn{3}{@{}c@{}}{ResNet} & \multicolumn{4}{@{}c@{}@{}}{ResNeSt} \\
                \cmidrule{2-4}\cmidrule{5-8}
                & 50 & 101 & 152 & 50 & 101 & 200 & 269 \\
                \midrule
                train & 95.86 & 96.35 & 96.58 & 97.46 & 97.96 & 98.49 & 0.9861 \\
                val & 93.52 & 93.33 & 93.95 & 95.44 & 96.38 & 96.2 & 96.31 \\
                \botrule
            \end{tabular*}
        \end{minipage}
    \end{center}
\end{table}

\begin{table}[!ht]
    \begin{center}
        \begin{minipage}{\textwidth}
        \caption{mIoU on Pseudo Segmentation Labels}\label{cams_vs_deeplabcam}%
        \begin{tabular*}{\textwidth}{@{\extracolsep{\fill}}cccccccc@{\extracolsep{\fill}}}
        \toprule%
        & \multicolumn{3}{@{}c@{}}{ResNet} & \multicolumn{4}{@{}c@{}}{ResNeSt} \\\cmidrule{2-4}\cmidrule{5-8}%
         & 50 & 101 & 152 & 50 & 101 & 200 & 269 \\
        \midrule
        CAM\footnotemark[1] & 48.93 & 50.33 & 50.83 & 53.49 & \textbf{57.78} & 57.33 & 54.75 \\
        DeepLabCAM\footnotemark[1] & 54.89 & 57.11 & 58.35 & 56.91 & \textbf{63.09} & 62.48 & 61.11 \\
        \botrule
        \end{tabular*}
        \footnotetext{This table contains the mIoU metric for the trainaug split of PASCAL VOC 2012 dataset}
        \end{minipage}
    \end{center}
\end{table}

The results of the CAM and DeepLabCAM versions are reported in Table \ref{cams_vs_deeplabcam}. Moreover, significant performance improvement is apparent for each backbone from that table as the gap is more than 3\% even in the closest form, rising higher than 7\%.  

The proposed method is proved to perform better than CAMs, with results presented in Table \ref{cams_vs_deeplabcam}, at least for the modified versions of the DeepLabv3. The next set of experiments is done with another segmentation model called UNet.

\subsubsection{CAM vs. UNetCAM}

In the second set of configurations, another well-known segmentation model with different characteristics and internals than DeepLab is hired called UNet.There are two different versions of the UNet composed in this experiment set, and they are named UNetCAM Learnable and UNetCAM Non-Learnable, as described in Section \ref{unetcam_description}. In the architecture of UNetCAM Learnable, the upsampling part of the decoder is constituted of transposed convolution layers, and for the other one, upsampling is done with bilinear interpolation. ResNet50 is selected to be the encoder part of the UNetCAM.

\begin{table}[!ht]
    \begin{center}
        \begin{minipage}{\textwidth}
            \caption{CAM vs UNetCAM}\label{cam_vs_unetcam_trainaug}%
            \begin{tabular*}{\textwidth}{@{\extracolsep{\fill}}cccc@{\extracolsep{\fill}}}
            \toprule%
             & CAM & UNetCAM Learnable & UNetCAM Non-Learnable \\
            \cmidrule{1-4}%
            train & 48.93 & 54.30 & 53.75 \\
            val & 47.59 & 53.01 & 52.41 \\
            \botrule
            \end{tabular*}
            \footnotetext{All the experiments are done by using trainaug split }
        \end{minipage}
    \end{center}
\end{table}

Table \ref{cam_vs_unetcam_trainaug} presents the results of both UNetCAM networks and CAM, and it shows that UNetCAM enhances the results by at least 3\%, which supports the experimentation results on DeepLabCAM.

\begin{table}[!ht]
    \begin{center}
        \begin{minipage}{\textwidth}
            \caption{CAM vs UNetCAM}\label{cam_vs_unetcam_train}%
            \begin{tabular*}{\textwidth}{@{\extracolsep{\fill}}cccc@{\extracolsep{\fill}}}
            \toprule%
             & CAM & UNetCAM Learnable & UNetCAM Non-Learnable \\
            train & 48.14 & 52.75 & 51.90 \\
            val & 46.60 & 51.32 & 50.56  \\
            \botrule
            \end{tabular*}
            \footnotetext{All the experiments are done by using train split }
        \end{minipage}
    \end{center}
\end{table}

To go a little further, we changed the data source with the current settings and used the train split instead of trainaug. Table \ref{cam_vs_unetcam_train} proves that the UNetCAM performs better even with fewer training images.

One another exciting examination is arranged to interpret the effects of the usage of dCRF. For this purpose, in Table \ref{crf_cam_vs_unetcam}, the metrics are calculated after dCRF is applied to the CAMs. We are investigating if our proposed method embeds the dCRF into the network and if the further application of the dCRF becomes somehow useless. The embedding idea matches with the results of Table \ref{crf_cam_vs_unetcam}, but it also shows that dCRF still has room to enhance the UNetCAM results.

\begin{table}[!ht]
    \begin{center}
        \begin{minipage}{\textwidth}
            \caption{CAM + dCRF vs UNetCAM + dCRF}\label{crf_cam_vs_unetcam}%
            \begin{tabular*}{\textwidth}{@{\extracolsep{\fill}}cccc@{\extracolsep{\fill}}}
            \toprule%
             & CAM & UNetCAM Learnable & UNetCAM Non-Learnable \\
            train & 53.18 & \textbf{59.90} & 58.74 \\
            val & 50.72 & \textbf{57.11} & 57.01  \\
            \botrule
            \end{tabular*}
            \footnotetext{All the experiments are done by using trainaug split }
        \end{minipage}
    \end{center}
\end{table}

\subsubsection{Investigation on the Iteration Procedure}

The conception of an iterative self-improved training procedure has its challenges, such as stabilization and optimization. In our proposal, the classification performance directly impacts the segmentation network as it uses the classification results. Furthermore, the segmentation network implicitly affects the features of the classifier network. If things go wrong for one task, they will go the same in the other, and the worsening will continue growing. The problem with optimization is that the overall network can be stuck in early steps without improving its performance. 

\begin{figure}[!ht]
    \centering
    \includegraphics[width=.8\linewidth]{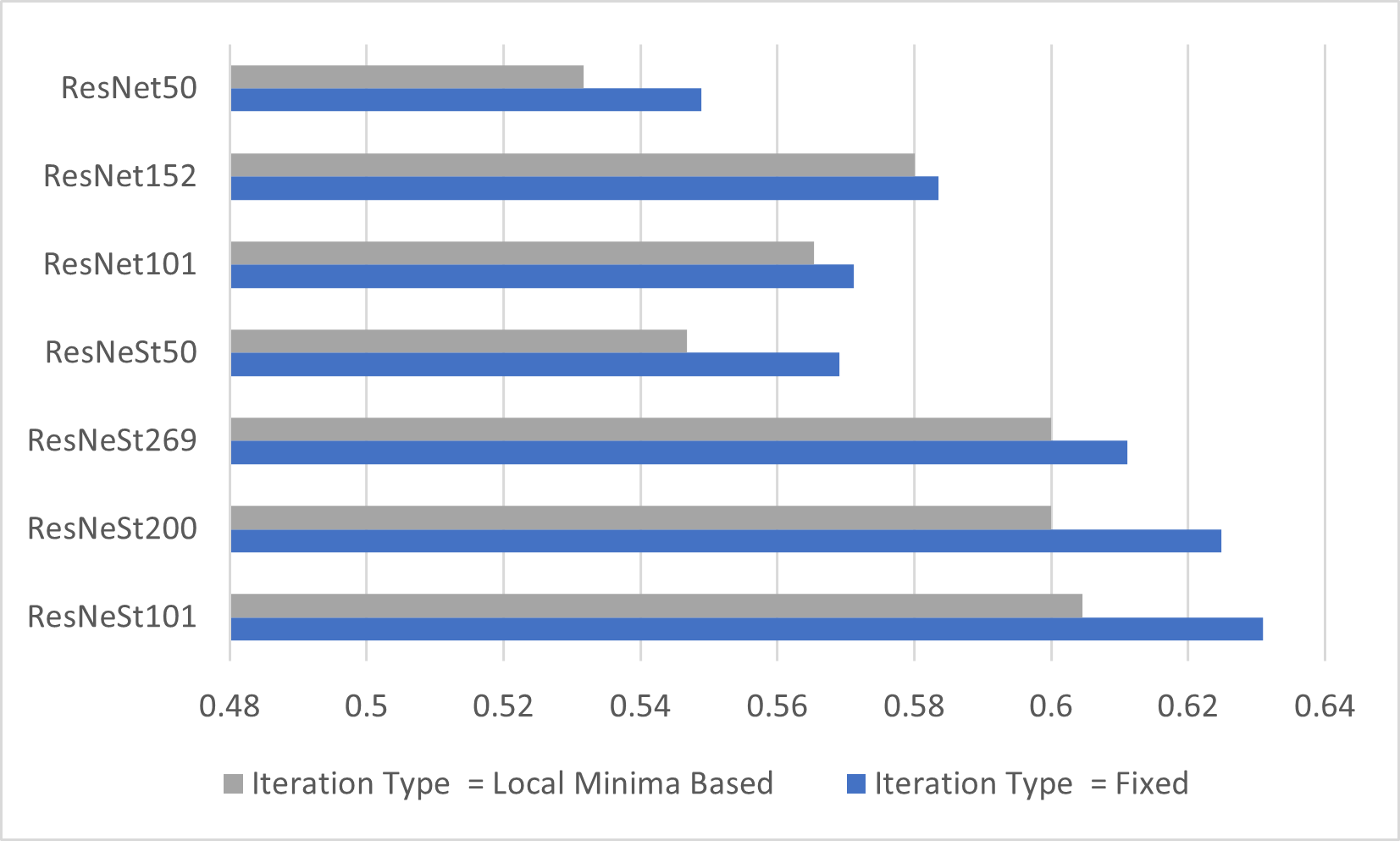}
    \caption{mIoU According to Iteration Procedure}
    \label{label_generation}
\end{figure}

Intuitively, these two dilemmas may happen in two different ways, (i) when the classifier and the segmentation part cannot find the room to optimize themselves,  (ii) when one of the tasks dominates the other in the training phase. To examine these potential problems, we set experimentations by changing the frequency of pseudo-ground-truth creation. The experiments focus on two main configurations; the first is waiting until the losses converge, and the second is fixing the generation frequency manually at the beginning of the training procedure.

Figure \ref{label_generation} represents the experimental results on the frequency of the pseudo-ground-truth generation, and from these results, it is clear that fixing the frequency performs better. Intuitively, this makes sense as this is an instance of the expectation-maximization problem, and giving room to each task and training in relaxed conditions performs better than strict rules.

\subsubsection{Investigation on Image Resolutions}

There are several differences between the applications of the segmentation and classification problems. One of them is that their input resolution preferences are different. Most of the classification architecture in their original forms accept the images whose resolutions vary between 224 to 299. However, it increases to 512 or higher for the popular segmentation architectures with viable reasons. As the proposed network supports segmentation and classification simultaneously, studying the different input sizes makes sense to reveal the framework's stability and robustness.

In this part of the study, the experimentation reveals the effects of the input resolutions on the proposed method, and four input sizes are selected to run tests 224x224, 320x320, 448x448, and 512x512.

\begin{table}[!ht]
    \begin{center}
    \begin{minipage}{\textwidth}
    \caption{Experiments with Different Input Sizes}\label{resolutions}%
    \begin{tabular*}{\textwidth}{@{\extracolsep{\fill}}ccccc@{\extracolsep{\fill}}}
    \toprule%
    & 256$\times$256 & 320$\times$320 & 448$\times$448 & 512$\times$512 \\
    \midrule
    CAM & 57.52 & 57.78 & 55.34 & 54.94 \\
    DeepLabCAM & 62.86 & 63.09 & 57.94 & 56.26 \\
    
    \botrule
    \end{tabular*}
    \footnotetext{This table contains the mIoU metric between predicted segmentation labels and original segmentation labels }
    \end{minipage}
    \end{center}
\end{table}

According to the reported Table \ref{resolutions}, the proposed architecture performs better than the original CAM for each input size. In addition to that, when the input size is 320x320, it reaches its best.

These four experiments prove that the proposed network performs better than CAMs when employed as segmentation labels for the dataset. It engages the presented network results as affinity labels and executes the pipeline to achieve the final segmentation model.

\subsubsection{Qualitative Results}

\begin{figure}[!ht]
    \centering
    \includegraphics[width=1.0\columnwidth]{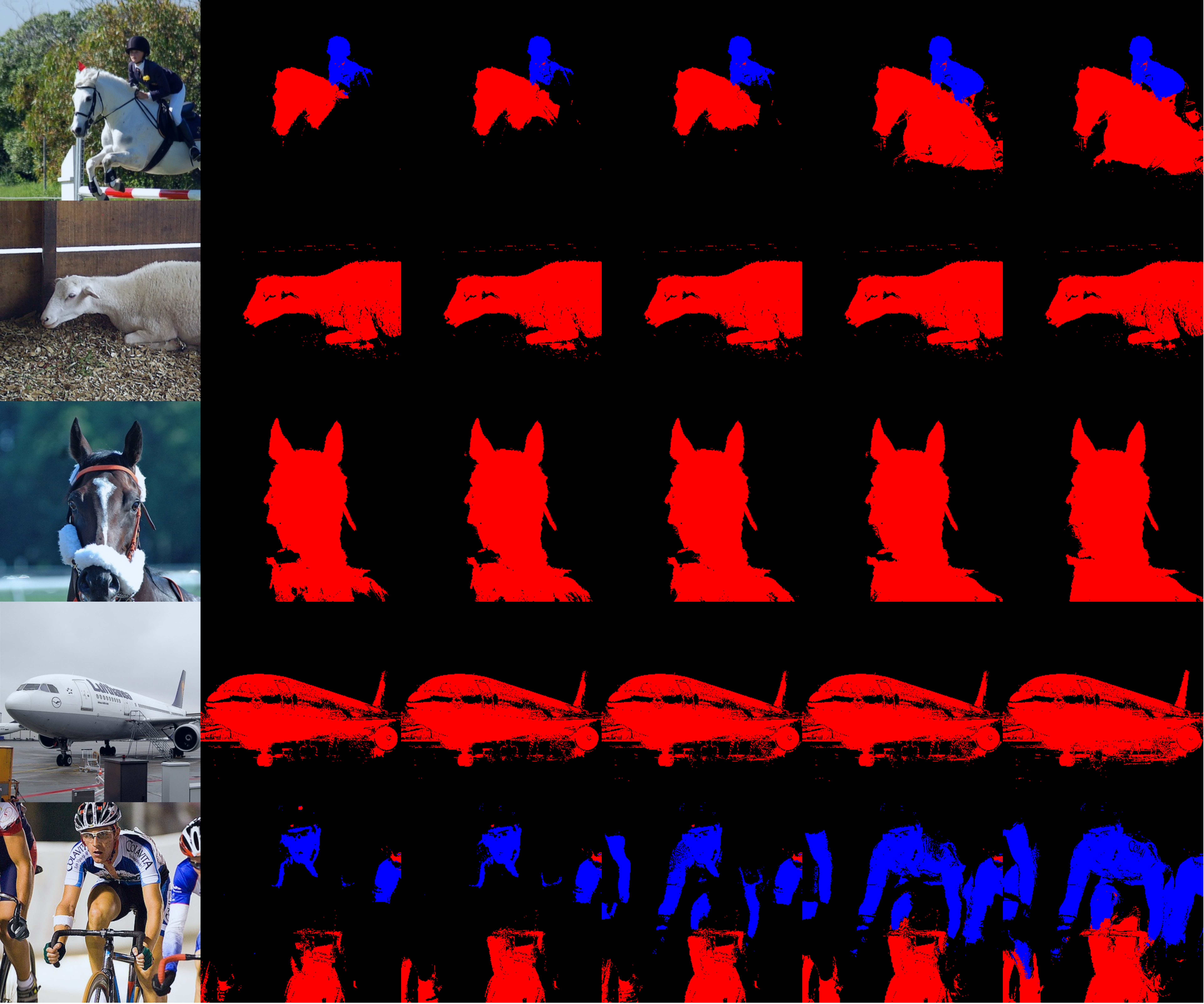}
    \caption{Visualization of Pseudo Segmentation Labels}
    \label{pseudo_labels}
\end{figure}

Qualitative investigations can be done on the visualizations in Figure \ref{pseudo_labels} to understand the effects of the segmentation loss and iterations more clearly. The first column of the figure shows the original image, and the other columns from left to right contain the pseudo-segmentation labels generated after epochs 5, 15, 25, 35, and 45, respectively. For the images in the first and the last rows, it is evident that the latter pseudo masks are more accurate than the previous ones. However, in the fourth row, it first begins to perform better, and in the last column, some pixels that are part of the background start to be marked as a plane, which is evidence of becoming poorer. Even though it is hard to say the effects of the framework are in the positive or negative direction, it is obvious from the images; one thing is clear, the proposed model has effects on the qualitative results. 

\subsubsection{AffinityNet Training}

AffinityNet is the network proposed in PSA to learn class-agnostic pixel-wise affinities. The model is trained on the images with a target called the affinity labels, generated from the denseCRF algorithm on CAMs of the DeepLabCAM network. There are two threshold values to apply on CAMs to generate affinity labels more confidently, one for the background and the other for the foreground, with the values 0.1 and 0.3, respectively. The foreground threshold has the same value as the threshold used in DeepLabCAM to optimize according to it. We followed the same steps to train and evaluate the model as performed in the original study. In addition to the overall procedure, we only adjust the backbone instead of ResNet-38 in our experiments, and we use deeper versions and versions of ResNeSt architecture. As the optimizer, the Poly optimizer is chosen. Moreover, AffinityNet trained for three epochs with batch size 16 on 512×512 resolution images.

In the end-to-end pipeline, the backbones are held in common generally. It means that we operate the same backbone in DeepLabCAM, AffinityNet, and the Segmentation model unless otherwise declared.

\begin{table}[!ht]
    \begin{center}
    \begin{minipage}{\textwidth}
    \caption{mIoU on Pseudo Segmentation Labels}\label{miou_of_cams}%
    \begin{tabular*}{\textwidth}{@{\extracolsep{\fill}}cccccccc@{\extracolsep{\fill}}}
    \toprule%
    & \multicolumn{3}{@{}c@{}}{ResNet} & \multicolumn{4}{@{}c@{}}{ResNeSt} \\\cmidrule{2-4}\cmidrule{5-8}%
    & 50 & 101 & 152 & 50 & 101 & 200 & 269 \\
    \midrule
    CAM\footnotemark[1] & 48.93 & 50.33 & 50.83 & 53.49 & \textbf{57.78} & 57.33 & 54.75 \\
    DeepLabCAM\footnotemark[1] & 54.89 & 57.11 & 58.35 & 56.91 & \textbf{63.09} & 62.48 & 61.11 \\
    DeepLabCAM+RW\footnotemark[1] & 74.25 & 76.40 & 78.54 & 75.65 & 76.78 & 77.47 & \textbf{79.10} \\
    DeepLabCAM+RW+CRF\footnotemark[1] & 74.83 & 77.09 & 79.41 & 75.97 & 77.29 & 77.93 & \textbf{79.64} \\
    \botrule
    \end{tabular*}
    \footnotetext{This table contains the mIoU metric between CAMs and original segmentation labels for the trainaug split of Pascal VOC2012 }
    \end{minipage}
    \end{center}
\end{table}
Table \ref{miou_of_cams} presents the mIoU values of several backbones with different settings. The first two rows contain the results of CAM and DeepLabCAM, respectively. The third row shows the improvements done on DeepLabCAM results by applying Random Walk with AffinityNet predictions. In addition to these, the fourth row also has one more step employing denseCRF. The results in table \ref{miou_of_cams} prove that AffinityNet performs better when it has more accurate affinity labels and also deeper networks achieve better performances.

The comparison between our proposed method with AffinityNet and previous studies with AffinityNet is handled in Section \ref{comparison_sota}.

\subsubsection{Segmentation model training}

According to the previous studies, three different configurations are organized on the version of the DeepLab family to make fair comparisons. DeepLabv2 architecture with the ResNet-101 backbone is exercised in the first set of experiments. The second set of experiments is done by upgrading the DeepLab version to 3 and keeping the same backbone. The last experiments are performed with DeepLabv3 again but with various backbones. For all of them, the input image resolution is 512×512, and the training is stopped after 50 epochs.

\begin{table}[!ht]
    \begin{center}
    \begin{minipage}{\textwidth}
    \caption{mIoU on Segmentation Labels with DeepLabv3}\label{miou_on_seg_deeplabv3}%
    \begin{tabular*}{\textwidth}{@{\extracolsep{\fill}}cccccccc@{\extracolsep{\fill}}}
    \toprule%
    & \multicolumn{3}{@{}c@{}}{ResNet} & \multicolumn{4}{@{}c@{}}{ResNeSt} \\\cmidrule{2-4}\cmidrule{5-8}%
     & 50 & 101 & 152 & 50 & 101 & 200 & 269 \\
    \midrule
    Val & 67.87 & 70.51 & 72.28 & 69.77 & 73.52 & \textbf{74.90} & 74.76 \\
    Test & 67.89 & 71.45 & 72.41 & 69.88 & 73.25 & \textbf{74.98} & 74.90 \\
    \botrule
    \end{tabular*}
    \footnotetext{This table contains the mIoU metric between predicted segmentation labels and original segmentation labels }
    \end{minipage}
    \end{center}
\end{table}

The results of the segmentation model with different backbones are reported in Table \ref{miou_on_seg_deeplabv3}. In this Table, the backbone of all three models, DeepLabCAM, AffinityNet, and DeepLabv3, is kept precisely the same in the end-to-end pipeline.  

\begin{table}[!ht]
    \begin{center}
    \begin{minipage}{\textwidth}
    \caption{mIoU on Segmentation Labels with DeepLabv3 (ResNet101)}\label{miou_on_seg_deeplabv3_101}%
    \begin{tabular*}{\textwidth}{@{\extracolsep{\fill}}cccccccc@{\extracolsep{\fill}}}
    \toprule%
    & \multicolumn{3}{@{}c@{}}{ResNet} & \multicolumn{4}{@{}c@{}}{ResNeSt} \\\cmidrule{2-4}\cmidrule{5-8}%
     & 50 & 101 & 152 & 50 & 101 & 200 & 269 \\
    \midrule
    Val & 69.09 & 70.51 & 71.54 & 70.39 & 73.55 & \textbf{73.70	} & 73.37 \\
    Test & 69.46 & 71.45 & 71.62 & 70.43 & 73.39 & \textbf{74.50} & 73.50 \\
    \botrule
    \end{tabular*}
    \footnotetext{This table contains the mIoU metric between predicted segmentation labels and original segmentation labels }
    \end{minipage}
    \end{center}
\end{table}

In Table \ref{miou_on_seg_deeplabv3_101}, the backbone of the last segmentation model is kept the same as ResNet-101 as it is widely used in the previous experiments.

\begin{table}[!ht]
    \begin{center}
    \begin{minipage}{\textwidth}
    \caption{mIoU on Segmentation Labels with DeepLabv2 (ResNet101)}\label{miou_on_seg_deeplabv2_101}%
    \begin{tabular*}{\textwidth}{@{\extracolsep{\fill}}ccccc@{\extracolsep{\fill}}}
    \toprule%
    & ResNet-50 & ResNet-101 & ResNeSt-200 \\
    \midrule
    Val & 68.64 & 70.60 & \textbf{72.60	} \\
    Test & 68.83 & 70.38 & \textbf{72.94} \\
    \botrule
    \end{tabular*}
    \footnotetext{This table contains the mIoU metric between predicted segmentation labels and original segmentation labels }
    \end{minipage}
    \end{center}
\end{table}

In Table \ref{miou_on_seg_deeplabv2_101}, to compare our results with the previous studies, we train the DeepLabv2 segmentation model with the pseudo-segmentation labels obtained from ResNet-50, ResNet-101 backbones.

\subsubsection{Comparison with SOTA}\label{comparison_sota}

In Table \ref{sota}, we compare our best results with the previous studies, which are state-of-the-art at their time. 

\begin{table}[!ht]
\begin{center}
\begin{minipage}{\textwidth}
\caption{Comparison of the results with literature}\label{sota}%
\begin{tabular*}{\textwidth}{@{\extracolsep{\fill}}cccccc@{\extracolsep{\fill}}}
\toprule%
Method & Backbone  & CAM\footnotemark[1] & Pseudo Masks\footnotemark[1] & Seg Masks\footnotemark[2] & Seg Masks\footnotemark[3] \\
\midrule
PSA & ResNet-38  & 48 & 59,7 & 61,7 & 63,7 \\
IRNet & ResNet-50  & 48,3 & 65,9 & 63,5 & 64,8 \\
FickleNet & ResNet-101  & - & - & 64.9 & 65.3\\ 
ICD & ResNet-101  & - & - & 64.1 & 64.3\\
SEAM & ResNet-38  & 55.4 & 63.4 & 64.5 & 65.7\\
CDA & ResNet-38 & 58.4 & 66.4 & 66.1 & 66.8 \\
PuzzleCAM & ResNeSt-101 & 61.85 & 72.46 & 66.9 & 67.7 \\
WSGCN & ResNet-101 & - & - & 68.7 & 69.3 \\
Ours\footnotemark[4] & ResNet-101 & 57,11 & 77.09 & 70.61 & 70.38 \\
DRS v1\footnotemark[4] & ResNet-101 &  - & - & 70.4 & 70.7 \\
EPS\footnotemark[4] & ResNet-101 & 69.4 & 71.6 & 70.9  & 70.8 \\
DRS v2\footnotemark[4] & ResNet-101 & - & - & 71.2 & 71.4 \\
Ours\footnotemark[5] & ResNet-101 & 57,11 & 77.09 & 70.51 & 71.45 \\
SPML\footnotemark[4] & ResNet-101 & - & - & 69.5 & 71.6 \\
EPS\footnotemark[7] & ResNet-101 & 69.4 & 71.6 & 71.0  & 71.8 \\
PuzzleCAM & ResNeSt-269 & 62.45 & 74.67 & 71.9  & 72.2 \\
Ours\footnotemark[5] & ResNeSt-200 & 62.48 & 77.93 & \textbf{72.60} & \textbf{72.94} \\
Ours\footnotemark[6] & ResNeSt-200 & 62.48 & 77.93 & \textbf{74.90} & \textbf{74.98} \\
\botrule
\end{tabular*}
\footnotetext{This table contains the mIoU metric between predicted segmentation labels and original segmentation labels }
\footnotetext[1]{The mIoU metric calculated on train split}
\footnotetext[2]{The mIoU metric calculated on val split}
\footnotetext[3]{The mIoU metric calculated on test split}
\footnotetext[4]{DeepLabv2 with backbone ResNet-101 used as Segmentation Model}
\footnotetext[5]{DeepLabv3 with backbone ResNet-101 used as Segmentation Model}
\footnotetext[6]{DeepLabv3 with backbone ResNeSt-200 used as Segmentation Model}
\footnotetext[7]{DeepLabv1 with backbone ResNet-101 used as Segmentation Model}
\end{minipage}
\end{center}
\end{table}

Table \ref{sota} can be interpreted like this, in overall performance, the best model is our proposed model with ResNeSt-269 backbone without using any additional data or saliency map with DeepLabv3 with a margin over \%2.5. When the backbone is kept the same between the studies as ResNet-101, the best performance is observed in the EPS study, which uses extra saliency map data in the training phase. When there is a restriction to not use any additional data in the training phase, our model becomes the second-best one with \%70.38 mIoU with a small margin on SPML which also does not use any additional information. In addition, EPS and DRS studies use saliency map data as the input in the other results in the table.


\section{Conclusion}\label{conclusion}
In this study, we propose a novel framework for WSSS. It employs an iterative self-improved approach in an encoder-decoder-based segmentation model to obtain coherent CAMs with segmentation labels. The proposed framework contains various elements from the available techniques in the literature. Therefore, it tries to combine the strengths of those methods.     

The extensive experiments show that the proposed framework is challenged and proven to improve the CAMs with quantitative and qualitative results. The iterative approach is optimized with the help of a simple modification in the well-known loss function, Cross-Entropy, to achieve the goal. In addition, a specific implementation of the proposed framework achieves the state-of-the-art Pascal VOC 2012 dataset.

As a possible future work, in theory, there is room to improve the loss function to stabilize the training phase of the proposed framework more. And also, combining AffinityNet with the proposed network will significantly affect the results as AffinityNet learns from pixel-level information from the images, and adding more pixel-level loss to the proposed network may improve the results.

\backmatter


\bibliography{sn-bibliography}



\end{document}